\documentclass[conference]{IEEEtran}
\IEEEoverridecommandlockouts

\usepackage{booktabs} 
\usepackage{multirow}

\usepackage[bottom]{footmisc}
\usepackage{blindtext}
\usepackage{graphicx}
\usepackage[font=footnotesize]{caption,subcaption}
\usepackage{wrapfig}

\usepackage{flushend}
\usepackage{amsmath,amsfonts,amsthm,bm}
\usepackage{stfloats}
\usepackage{graphicx}
\usepackage{textcomp}
\usepackage{xcolor}
\usepackage{booktabs} 

\usepackage{amsmath}
\usepackage{rotating}
\usepackage{booktabs}
\usepackage{listings}

\usepackage{listings}

\usepackage{ragged2e} 
\usepackage{algorithm}
\usepackage[algo2e]{algorithm2e} 
\usepackage{algpseudocode}

\def\BibTeX{{\rm B\kern-.05em{\sc i\kern-.025em b}\kern-.08em
    T\kern-.1667em\lower.7ex\hbox{E}\kern-.125emX}}
\begin{document}

\title{Quantized and Interpretable Learning Scheme for Deep Neural Networks in Classification Task\\
}

\author{\IEEEauthorblockN{Alireza Maleki}
\IEEEauthorblockA{\textit{Department of Technology Management} \\
\textit{University of Tehran}\\
Tehran, Iran \\
alireza.maleki@alum.sharif.edu}
\and
\IEEEauthorblockN{Mahsa Lavaei}
\IEEEauthorblockA{\textit{Electrical and Computer Engineering} \\
\textit{University of Tehran}\\
Tehran, Iran \\
M.lavaei@ut.ac.ir}
\and
\IEEEauthorblockN{Mohsen Bagheritabar}
\IEEEauthorblockA{\textit{Electrical Engineering Department} \\
\textit{, University of Cincinnati}\\
Ohio, USA  \\
baghermn@mail.uc.edu}
\and
\IEEEauthorblockN{Salar Beigzad}
\IEEEauthorblockA{\textit{ Software and Engineering} \\
\textit{University of St. Thomas, Minnesota}\\
Minnesota, USA \\
beig2558@stthomad.edu}
\and
\IEEEauthorblockN{Zahra Abadi}
\IEEEauthorblockA{\textit{Electrical Engineering Department} \\
\textit{Tehran University}\\
Tehran, Iran  \\
zkarkehabadi79@gmail.com}}

\maketitle

\begin{abstract}
Deep learning techniques have proven highly effective in image classification, but their deployment in resource-constrained environments remains challenging due to high computational demands. Furthermore, their interpretability is of high importance which demands even more available resources. In this work, we introduce an approach that combines saliency-guided training with quantization techniques to create an interpretable and resource-efficient model without compromising accuracy. We utilize Parameterized Clipping Activation (PACT) to perform quantization-aware training, specifically targeting activations and weights to optimize precision while minimizing resource usage. Concurrently, saliency-guided training is employed to enhance interpretability by iteratively masking features with low gradient values, leading to more focused and meaningful saliency maps. This training procedure helps in mitigating noisy gradients and yields models that provide clearer, more interpretable insights into their decision-making processes. To evaluate the impact of our approach, we conduct experiments using famous Convolutional Neural Networks (CNN) architecture on the MNIST and CIFAR-10 benchmark datasets as two popular datasets. We compare the saliency maps generated by standard and quantized models to assess the influence of quantization on both interpretability and classification accuracy. Our results demonstrate that the combined use of saliency-guided training and PACT-based quantization not only maintains classification performance but also produces models that are significantly more efficient and interpretable, making them suitable for deployment in resource-limited settings.
\end{abstract}


Deep Neural Networks (DNNs) have become a cornerstone in advancing fields such as computer vision, natural language processing, autonomous driving, and healthcare. By extracting complex patterns from large-scale datasets, they have achieved unprecedented accuracy in tasks like image recognition, language translation, and predictive analytics, redefining the capabilities of modern technology. Despite these successes, DNNs are often criticized for their "black-box" nature, which obscures their internal decision-making processes and creates challenges in understanding how predictions are made. This lack of transparency becomes particularly concerning in critical domains such as medicine, finance, and autonomous systems, where trust, reliability, and safety hinge on understanding a model’s rationale \cite{caruana2015intelligible, li2018tell, hassanpour2024overcoming, karkehabadi2024ffcl}. To address these issues, substantial research efforts have been dedicated to developing interpretability methods that explain DNN decision-making processes. One widely adopted approach is the generation of saliency maps, which visually highlight the input features most influential in a model's prediction. These maps, typically created using gradient-based techniques, calculate the gradients of model outputs with respect to input features, revealing the regions of importance \cite{selvaraju2017grad, shrikumar2017learning}. Saliency maps provide valuable insights into model behavior by illustrating what the model focuses on during decision-making. However, the reliability of these maps can be compromised by noise and other artifacts, leading to ambiguous or misleading interpretations \cite{singh2017hide, kindermans2016investigating}. To mitigate such issues, advanced methods like SmoothGrad have been proposed, which reduce noise by averaging multiple saliency maps generated from perturbed versions of the input data \cite{smilkov2017smoothgrad}. Additionally, techniques such as Integrated Gradients \cite{sundararajan2017axiomatic} and Layer-wise Relevance Propagation \cite{bach2015pixel} have improved the stability and robustness of saliency maps by refining the backpropagation process. Despite these advances, ensuring that saliency maps are both accurate and robust remains a challenge, particularly when the model or input data experiences slight variations \cite{adebayo2018sanity, ghorbani2019interpretation}. In parallel to the development of interpretability techniques, optimizing DNNs for deployment in resource-constrained environments has gained growing interest. Quantization, which reduces the precision of model parameters and activations, is one of the key methods for achieving efficient neural network deployment. 
Here’s a refined version of your text with improved flow and clarity:

Quantization reduces both memory usage and computational overhead by converting 32-bit floating-point representations to lower-precision formats, such as 8-bit integers. This transformation enables the deployment of complex models on resource-constrained devices, including mobile phones and embedded systems \cite{Weng2023}. It plays a critical role in executing deep neural networks (DNNs) efficiently on edge devices, where computational power, memory capacity, and energy availability are limited \cite{Hubara2017, pour2024urban}. Quantization-Aware Training (QAT) incorporates quantization into the training process, ensuring models retain high accuracy despite reduced precision. Unlike Post-Training Quantization (PTQ), which applies quantization after the model is fully trained, QAT enables the model to learn and adapt to lower-precision representations during training. This approach results in superior accuracy, particularly in environments with stringent resource constraints \cite{Nagel2019}. QAT is particularly advantageous for edge devices such as smartphones and IoT systems, where energy efficiency and computational speed are crucial. Specialized hardware accelerators, such as Field-Programmable Gate Arrays (FPGAs) and Application-Specific Integrated Circuits (ASICs), further enhance the performance of QAT-trained models by optimizing low-precision computations \cite{Han2016}. These hardware solutions enable real-time deployment of quantized models in applications requiring both precision and efficiency, such as autonomous vehicles and medical diagnostics. The synergy between QAT and advanced hardware accelerators facilitates the development of scalable, energy-efficient AI systems, delivering high performance even in resource-constrained settings.
This paper explores the intersection of these two domains—model interpretability and quantization—by conducting a comprehensive comparison between "regular model " and "quantized model" saliency-based training. Specifically, we investigate whether quantization influences the generation of saliency maps, and if so, how these effects manifest across different scenarios. By applying the saliency method to both high-precision and quantized neural networks, we analyze the resulting maps in terms of clarity, stability, and consistency with high-precision models. Our study aims to contribute to the ongoing discussion about balancing model efficiency with interpretability, providing insights that could inform the development of deployable yet transparent AI systems.

\section{Related Works}
Saliency maps are a widely used tool for interpreting DNNs by highlighting the input features that most influence the model's predictions. Techniques like Grad-CAM \cite{selvaraju2017grad} and Integrated Gradients \cite{sundararajan2017axiomatic} are popular methods for generating these visual explanations. However, these methods often suffer from noise and instability, leading to challenges in accurately interpreting model behavior \cite{smilkov2017smoothgrad, adebayo2018sanity}. To mitigate these issues, approaches such as SmoothGrad \cite{smilkov2017smoothgrad} and Layer-wise Relevance Propagation (LRP) \cite{bach2015pixel} have been developed, focusing on enhancing the robustness and clarity of saliency maps. Saliency Guided Training (SGT) has emerged as a promising approach to address the limitations of traditional saliency methods. Unlike post-hoc techniques, which generate explanations after the model has been trained, SGT incorporates interpretability into the training process itself. By iteratively masking input features with low gradient values and training the model to maintain consistent outputs, SGT reduces noise in saliency maps and improves their reliability. This approach not only enhances the interpretability of the model but also leads to better generalization, as demonstrated by Ismail et al. \cite{ismail2021improving}. SGT has demonstrated a remarkable ability to enhance the comprehensiveness and stability of saliency maps across diverse neural architectures, including Convolutional Neural Networks (CNNs), Recurrent Neural Networks (RNNs), and Transformers. Simultaneously, advancements in neural network quantization have facilitated the efficient deployment of deep neural networks (DNNs) in resource-constrained environments by optimizing model precision and reducing computational demands \cite{Weng2023,ismail2021improving}. Techniques like Quantization-Aware Training \cite{Nagel2019} and mixed-precision quantization \cite{Choi2018} have been crucial in maintaining model accuracy despite reduced computational complexity. Parameterized Clipping Activation (PACT), introduced by Choi et al. \cite{Choi2018}, is a notable advancement in quantization techniques. PACT optimizes the clipping level of activations during training, allowing for aggressive quantization without significant loss in model accuracy. This method has demonstrated that both weights and activations can be quantized to low precision while still achieving accuracy comparable to full-precision models. Despite these advancements, the impact of quantization on the interpretability of models through saliency maps has not been explored well. Most prior work on saliency methods has focused on full-precision models, while the interpretability of quantized models remains an open question. This paper aims to bridge this gap by comparing regular and quantized saliency-based models to assess how precision reduction affects the generation and reliability of these interpretability tools.

\section{Interpretability in Machine Learning}

Interpretability in machine learning is essential for making complex model decisions understandable to humans, which is particularly important for ensuring that AI systems are transparent and trustworthy. Despite the remarkable accuracy of deep learning models, their opaque "black-box" nature limits our ability to justify predictions, especially in sensitive applications such as healthcare, finance, and autonomous systems\cite{karkehabadihlgm,bakhshi2024novel}. Interpretability techniques aim to clarify the internal workings of these models, bridging the gap between complex algorithmic processes and human decision-making, thereby enhancing the reliability of AI systems \cite{chakraborty2017interpretability, karkehabadi2024smoot}. One influential approach to improving interpretability is the "grafting" technique introduced by \cite{perkins2003grafting}, which incorporates feature selection directly into the gradient descent process. This method iteratively selects and integrates features during optimization, making it efficient and adaptable to both linear and non-linear models in classification and regression tasks. Grafting has served as a foundation for other interpretability methods, such as stochastic gradient descent-based soft decision trees \cite{frosst2017distilling}. Additional methods have emerged to enhance interpretability further. Tools like LIME (Local Interpretable Model-agnostic Explanations) provide insights into model behavior by revealing the implicit rules behind predictions. However, these methods often face scalability challenges when applied to large datasets or complex models. To overcome these issues, \cite{ross2017right} introduced a technique that penalizes input gradients, refining decision boundaries and improving interpretability by producing clearer explanations of model behavior.

\section{Saliency-Guided Training}

Saliency-guided approaches focus on identifying and highlighting the most relevant features in input data that influence a model’s predictions. Gradient-based techniques, such as vanilla Gradient \cite{simonyan2013deep}, Grad-CAM \cite{selvaraju2017grad}, and SmoothGrad \cite{smilkov2017smoothgrad}, generate saliency maps by computing the gradients of model outputs with respect to inputs, providing a visual representation of the key areas that affect the model's decisions.

While post-hoc methods offer valuable insights, they can sometimes produce noisy or unclear saliency maps, which may highlight irrelevant features. Saliency Guided Training (SGT) addresses these issues by incorporating saliency directly into the training process. SGT improves both accuracy and interpretability by teaching the model to focus on crucial features during training. By minimizing the Kullback-Leibler (KL) divergence between original and masked outputs, SGT encourages the model to ignore less important features and emphasize the most relevant ones, resulting in clearer, more reliable saliency maps \cite{ismail2021improving}.

\begin{algorithm}[H]
\begin{algorithmic}

 \State Training samples $X$, number of features to be masked $k$, learning rate $\tau$, hyperparameter $\lambda$, Initialize $f_{\theta}$\\{Pre-load and randomize for new training}\\
 \For{$i = 1$ \textbf{to} epochs}{
    \For{$i = 1$ \textbf{to} epochs}{
        \scriptsize \# {Obtain the sorted indices \( I \) based on the gradient of the output with respect to the input:} \normalsize
        \State $I = \text{$ S $}(\nabla_X f_{\theta_i}(X))$ 
        
        \scriptsize \# {Mask the bottom $k$ features} \normalsize
        \State $\widetilde{X} = M_k(I, X)$ 
        
        \scriptsize \# {Compute the loss function:} \normalsize
        \State $L_i = \mathcal{L}(f_{\theta_i}(X), y) + \lambda \mathcal{D_{KL}}  (f_{\theta_i}(X) \| f_{\theta_i}(\widetilde{X}))$
        
        \scriptsize \# {Use the gradient to update network parameters.} \normalsize
        \State $f_{\theta_{i+1}} = f_{\theta_i} - \tau \nabla_{\theta_i} L_i$ 
   }{
  }
 }
 \caption{Saliency Guided Training\cite{ismail2021improving}}
 \label{SGT Alg}
\end{algorithmic}
\end{algorithm}

\section{Quantization and Interpretability}

Parallel to interpretability advancements, quantization techniques such as Quantization-Aware Training (QAT) \cite{Nagel2019} and mixed-precision quantization \cite{Choi2018} have made it feasible to deploy deep neural networks in resource-constrained environments by reducing model precision. However, the interaction between quantization and interpretability, particularly regarding saliency maps, remains underexplored. While most research has focused on full-precision models, understanding how quantization affects the clarity and stability of saliency maps is crucial for ensuring that efficient, low-precision models remain interpretable.

\section{Saliency-Guided Training (SGT)}

Algorithm \ref{SGT Alg} illustrates the Saliency-Guided Training (SGT) process, where model parameters \( \theta \) are iteratively updated by leveraging the importance of input features as measured by their gradients. This approach allows the model to focus on the most relevant features during training, thus enhancing interpretability and boosting predictive accuracy. SGT incorporates two key components in its loss function, ensuring that the model learns to prioritize significant features:

\begin{itemize}
    \item \textbf{Cross-Entropy Loss}: This standard loss function captures the discrepancy between the model’s predicted output \( f_{\theta}(X_i) \) and the true label \( y_i \), driving the model toward more accurate predictions.
    \[
    \text{Cross-Entropy Loss:} \quad \mathcal{L}(f_{\theta}(X_i), y_i)
    \]
    \item \textbf{Kullback–Leibler (KL) Divergence}: To ensure the model generalizes well, this term measures the divergence between the model’s output on the original input \( X_i \) and a masked version \( \widetilde{X}_i \), where less important features are masked out. Minimizing this divergence forces the model to concentrate on salient features.
    \[
    D_{KL}(P\|Q) = \sum_{x \in X} P(x) \log\left(\frac{P(x)}{Q(x)}\right)
    \]
\end{itemize}

The overall loss function for SGT balances accuracy with a focus on salient features:

\[
\text{Overall Loss:} \quad \sum_{i=1}^{n} \left[\mathcal{L}(f_{\theta}(X_i), y_i) + \lambda \mathcal{D_{KL}}(f_{\theta}(X_i) \| f_{\theta}(\widetilde{X}_i))\right]
\]

This process updates model parameters to minimize the overall loss, guiding the model to focus on important features. SGT improves both accuracy and interpretability, making it useful for applications where transparency is critical.

\subsection{Quantization}

Quantization is a key optimization technique that reduces the precision of model parameters and activations, leading to substantial reductions in memory usage and computational demands. By minimizing resource requirements, quantization enables the efficient deployment of neural networks in resource-constrained environments.

\subsubsection{Quantization-Aware Training (QAT)}

QAT integrates quantization into the training process. Unlike Post-Training Quantization (PTQ), where quantization occurs after training, QAT allows the model to learn quantization constraints during training. Simulated quantization is applied during both forward and backward passes, using the following quantization function:

\[
w_{\text{out}} = \Delta \cdot \text{clamp}\left(0, N_{\text{levels}} - 1, \text{round}\left(\frac{w_{\text{float}}}{\Delta}\right) - z\right)
\]

Here, \( \Delta \) represents the step size, \( N_{\text{levels}} \) is the number of quantization levels, and \( z \) is the zero-point offset. The Straight-Through Estimator (STE) is used in the backward pass to approximate the gradients for the quantized weights:

\[
\delta_{\text{out}} = \delta_{\text{in}} \cdot I_{w_{\text{float}} \in (w_{\text{min}}, w_{\text{max}})}
\]

QAT significantly narrows the accuracy gap between floating-point and quantized models, especially at lower bitwidth.

\subsubsection{Parameterized Clipping Activation for QAT}

The \textbf{PACT} method introduces a learnable clipping parameter \( \alpha \) during training, addressing the challenges of quantizing activation functions by dynamically adjusting the clipping threshold. The clipping function is defined as:

\[
a_{\text{clip}} = \text{clip}(a, -\alpha, \alpha)
\]

Here, \( \alpha \) is a learnable parameter optimized along with the network weights, minimizing quantization error for activations. PACT allows activations and weights to be quantized to ultra-low precision (e.g., 4-bits) while maintaining near full-precision accuracy \cite{Choi2018}.

\section{Methodology}
In this work, we propose a framework combining \textbf{Saliency-Guided Training (SGT)} and \textbf{Parameterized Clipping Activation (PACT)} to develop resource-efficient and interpretable models. The proposed approach focuses on two key aspects: enhancing the model's interpretability and reducing computational complexity, while maintaining classification accuracy.

\begin{enumerate}
    \item \textbf{Saliency-Guided Training (SGT)}: 
    This technique leverages the gradients of the model's output with respect to the input to highlight important features. In each training step, features with the lowest gradients (i.e., the least important features) are masked, enabling the model to focus on salient regions of the input. This process improves the interpretability of the model by making the saliency maps more meaningful. Additionally, we penalize the divergence between the predictions on the original input and the masked input using Kullback-Leibler (KL) divergence, ensuring that the model remains robust to slight perturbations.

    \item \textbf{PACT-based Quantization}: 
    To achieve resource efficiency, we apply \textbf{Parameterized Clipping Activation (PACT)}, a quantization-aware training technique. PACT introduces a learnable parameter, $\alpha$, which controls the clipping level of the activation function. This enables the model to dynamically adjust the activation range during training to minimize the quantization error. The activations are then quantized to $k$-bit precision for efficient computation. PACT ensures that the model is optimized for quantized environments, making it suitable for deployment in resource-constrained settings.

\end{enumerate}

The proposed algorithm in Algorithm \ref{SGT pact Alg} presents a Saliency-Guided Training with PACT Quantization method. It starts by initializing the network $f_{\theta}$ and the clipping parameter $\alpha$ for PACT. For each epoch, the algorithm computes the saliency map by calculating the gradient of the network’s output with respect to the input, sorts the features, and masks the bottom $k$ least important features from the input. The loss function consists of the prediction error and a KL divergence term to ensure consistency between original and masked outputs. Both the network weights and the clipping parameter $\alpha$ are updated using backpropagation. This process continues for multiple epochs to refine both the network and quantization parameters.

\begin{algorithm}[H]
\begin{algorithmic}

 \State \textbf{Input:} Training samples $X$, number of features to be masked $k$, learning rate $\tau$, hyperparameter $\lambda$, initial parameter $\alpha$ for PACT quantization
\State \textbf{Initialize:} $f_{\theta}$ with random weights, clipping parameter $\alpha$\\
     \For{$i = 1$ \textbf{to} epochs}{
        \For{$i = 1$ \textbf{to} epochs}{
            
            \textcolor{blue}{Obtain the sorted indices \( I \) based on the gradient of the output with respect to the input:} \normalsize
            \State $I = \text{$ S $}(\nabla_X f_{\theta_i}(X))$ 
            
            \textcolor{blue}{{Mask bottom $k$ features of the original input.}} \normalsize
            \State $\widetilde{X} = M_k(I, X)$ 
            
            \textcolor{blue}{Compute the loss function.} \normalsize
            \[
             L_i = \mathcal{L}(y_{\text{orig}}, y) + \lambda \mathcal{D}_{KL}(y_{\text{orig}} \| y_{\text{masked}})
            \]
            
            {\textcolor{blue}{Use the gradient to update network parameters.}} \normalsize
            
            {\textcolor{blue}{Update clipping parameter $\alpha$ using the PACT loss:}} \normalsize
            
            \[f_{\theta_{i+1}} = f_{\theta_i} - \tau \nabla_{\theta_i} L_i\]
            \[
            \alpha_{i+1} = \alpha_i - \tau \nabla_{\alpha_i} L_{PACT}
            \]
   }{
  }
 }
 \caption{\textcolor{blue}{\textbf{Saliency-Guided Training PACT Quantization}}}
 \label{SGT pact Alg}
     
\end{algorithmic}
\end{algorithm}

\section{PACT Quantization Scheme}
The quantization is applied to activations in the forward pass using the following equation:

\[
y = PACT(x) = 
\begin{cases} 
0, & x < 0 \\
x, & 0 \leq x < \alpha \\
\alpha, & x \geq \alpha
\end{cases}
\]

The activations are quantized to $k$-bit precision using:

\[
y_q = \text{round}(y \cdot \frac{2^k - 1}{\alpha}) \cdot \frac{\alpha}{2^k - 1}
\]

The gradient for $\alpha$ is computed using the Straight-Through Estimator (STE) for backpropagation, ensuring that the model learns an optimal clipping level during training.

\subsection{Training Scheme}

We train our models on the MNIST and CIFAR-10 datasets using a combination of standard and quantized versions of ResNet-20 architecture. The model undergoes saliency-guided training during each epoch, and quantization is applied through PACT, ensuring the model is both interpretable and resource-efficient. To generate saliency maps, we utilized the Captum library \cite{kokhlikyan2020captum}, which provides a wide range of tools for model interpretability. The models were trained on an NVIDIA 1080Ti GPU.

\begin{table}[H]
\centering
\begin{tabular}{|c|c|c|}
\hline
\textbf{Hyperparameter}         & \textbf{MNIST}  & \textbf{CIFAR-10} \\
\hline
Learning Rate          & 0.1    & 0.01 \\
\hline
Epochs                 & 50     & 50 \\
\hline
Batch Size             & 128    & 64  \\
\hline
Quantization Bits ($k$)  & 8      & 4   \\
\hline
Clipping Parameter ($\alpha$) & Learnable  & Learnable  \\
\hline
Saliency Masking Ratio & 50\%  & 50\% \\
\hline
Regularization Term ($\lambda$) & 0.1  & 0.05 \\
\hline
\end{tabular}
\caption{Training Hyperparameters for Saliency-Guided Training with PACT Quantization and CIFAR-10 values}
\label{table:hyperparams}
\end{table}

\subsection{Datasets}

\textbf{MNIST:}
The MNIST dataset \cite{lecun2010mnist} is a cornerstone in machine learning research, comprising 70,000 grayscale images of handwritten digits. Each image, with dimensions of $28 \times 28$ pixels, belongs to one of 10 categories, representing the digits 0 to 9. The dataset is split into 60,000 images for training and 10,000 images for testing, making it an indispensable resource for benchmarking image classification models.

\textbf{CIFAR-10:}
The CIFAR-10 dataset \cite{krizhevsky2009learning} is a widely used dataset of 60,000 color images, each with dimensions of $32 \times 32$ pixels. The dataset includes 10 evenly distributed categories, representing everyday objects such as airplanes, cars, and birds. With 50,000 training images and 10,000 testing images, CIFAR-10 serves as a benchmark for evaluating the performance of image classification techniques.

\section{Experiments and Results}

In this section, we present the outcomes of our experiments comparing the accuracy, interpretability, and efficiency of the quantized models generated using saliency-guided training and PACT-based quantization.

\subsection{Accuracy Comparison}

Table \ref{table:accuracy} provides a comparison of classification accuracy for the datasets. We assess the performance of the standard models against the quantized counterparts, demonstrating that our approach retains high accuracy despite the reduction in computational complexity.

\begin{table}[h]
\centering
\caption{Accuracy Comparison}
\label{table:accuracy}
\begin{tabular}{|c|c|c|}
\hline
\textbf{Dataset} & \textbf{PACT (Accuracy\%)} & \textbf{Ours (Accuracy\%)} \\ \hline
MNIST            & 99.23                                & 99.35                                  \\ \hline
CIFAR-10         & 75.55                                & 76.12                               \\ \hline
\end{tabular}
\end{table}

\subsection{Dynamic $\alpha$ Optimization}

\begin{figure}[t]
    \centering
    \includegraphics[width=0.42\textwidth]{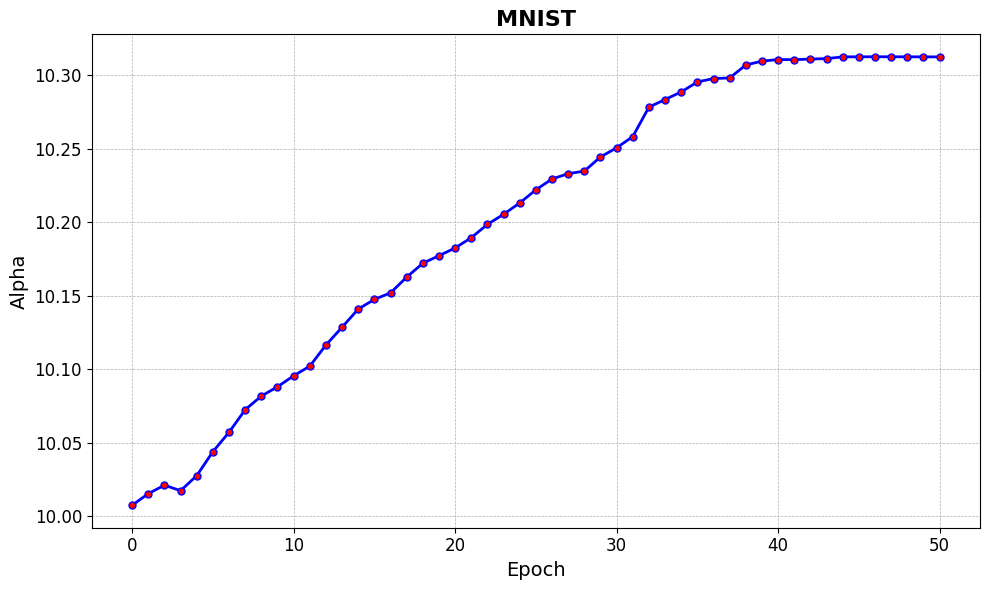}
    \caption{$\alpha$ Optimization for MNIST}
    \label{fig_alpha1}
\end{figure}

\begin{figure}[t]
    \centering
    \includegraphics[width=0.42\textwidth]{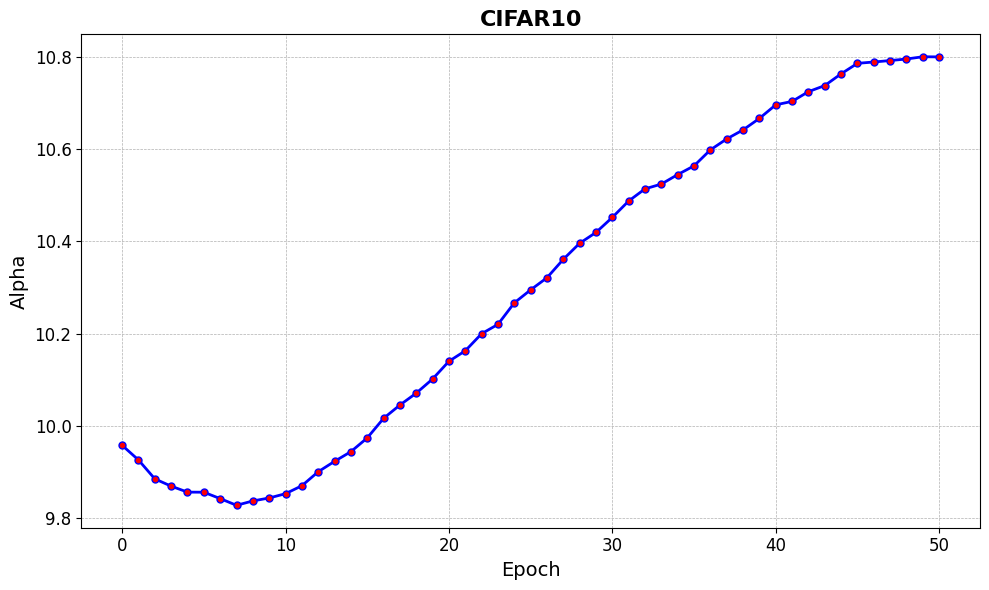}
    \caption{$\alpha$ Optimization for CIFAR10}
    \label{fig_alpha2}
\end{figure}

We propose a dynamic $\alpha$ optimization method to balance accuracy and resource efficiency during quantization-aware training. $\alpha$ controls the degree of quantization, initially set high for precision and gradually tuned to allow more aggressive quantization without significant accuracy loss. By updating $\alpha$ throughout training, the model finds the best value to balance the tradeoff between accuracy and quantization. Figure \ref{fig_alpha1} and \ref{fig_alpha2} illustrate how  $\alpha$ adjusts during training for both datasets.

\begin{figure}[ht!]
    \centering
    \includegraphics[width=0.42\textwidth]{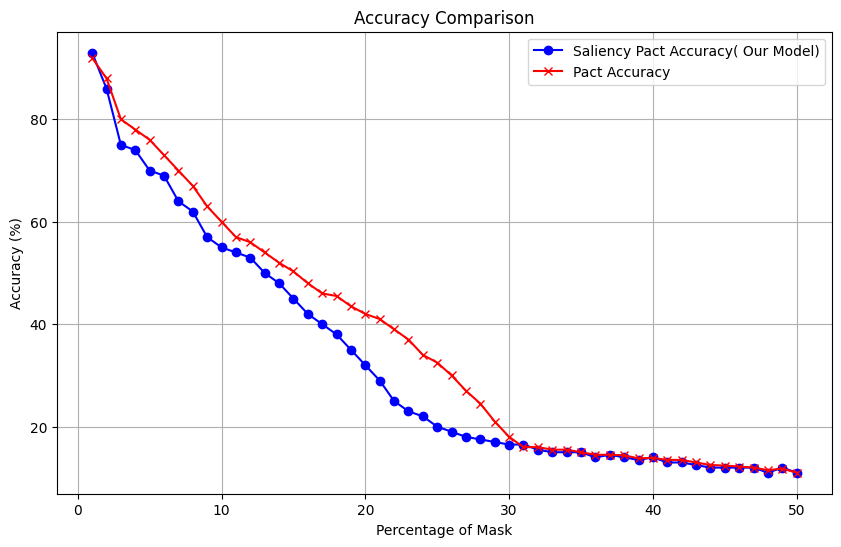}
    \caption{Accuracy drop comparison across different models on MNIST dataset with varying levels of masking percentages. It is shown that our approach's accuracy drops more as the masking percentage increases. This is an illustration that the model has learned more salient features during the training.}
    \label{fig:accuracy_drop}
\end{figure}

\subsection{Impact of Quantization on Model Accuracy}

We analyzed the effect of quantization on SGT models' accuracy by eliminating features based on saliency rankings. So models with lower bitwidths (higher quantization) experience sharper accuracy drops. This highlights the trade-off between aggressive quantization and the model’s ability to retain important features. Figure~\ref{fig:accuracy_drop} illustrates the effect of quantization on accuracy drop across different models

\section{Conclusion}
In this paper, we introduced a novel approach combining saliency-guided training with PACT-based quantization to develop efficient and interpretable deep learning models. Our method aims to tackle the challenge of deploying high-performance models in resource-constrained environments while maintaining robust classification accuracy. By incorporating saliency maps into the training process, we enhanced the interpretability of the models, allowing for more focused decision-making and clearer insights into their predictions. The experimental results on the MNIST and CIFAR-10 datasets demonstrate that the quantized models maintain near-equivalent accuracy to standard models, while their resource consumption is limited. Our results show that including saliency-based training can also improve the classification accuracy as well as increase the model's interpretability. Furthermore, to show the interpretability of the model, we have done an accuracy-drop evaluation on naive quantized and our approach. As Figure \ref{fig:accuracy_drop} suggests, our approach undergoes a steeper accuracy drop when a higher percentage of the input is masked. This shows that the model is more aware and interpretable considering the input. 
Our approach opens the door to deploying interpretable deep learning models in a variety of practical applications, especially in environments where computational resources are limited, such as edge devices and mobile platforms. Future work could extend this methodology to more complex datasets and architectures, further refining the balance between efficiency, accuracy, and interpretability.

\end{document}